\documentclass[a4paper, 10pt, conference]{ieeeconf}      
\usepackage{FG2024}
\usepackage{graphicx}
\usepackage{amsmath}
\usepackage{amsfonts}
\usepackage{amssymb}
\usepackage{multirow}
\usepackage{multicol}

\FGfinalcopy 

\IEEEoverridecommandlockouts                              
\def\FGPaperID{197} 

\title{\LARGE \bf
FE-Adapter: Adapting Image-based Emotion Classifiers to Videos
}

\author{\parbox{16cm}{\centering
    {\large Shreyank N Gowda$^1$, Boyan Gao$^1$ and David A. Clifton$^1$$^2$}\\
    {\normalsize
    $^1$ Department of Engineering Sciences, University of Oxford, OX3 7DQ Oxford, UK\\
    $^2$ Oxford Suzhou Centre for Advanced Research, University of Oxford, Suzhou 215123, Jiangsu, China}}
}

\usepackage{fancyhdr}
\begin{document}

\ifFGfinal
\thispagestyle{empty}
\pagestyle{empty}
\else
\author{Anonymous FG2024 submission\\ Paper ID \FGPaperID \\}
\pagestyle{plain}
\fi
\maketitle

\thispagestyle{fancy}

\begin{abstract}

Utilizing large pre-trained models for specific tasks has yielded impressive results. However, fully fine-tuning these increasingly large models is becoming prohibitively resource-intensive. This has led to a focus on more parameter-efficient transfer learning, primarily within the same modality. But this approach has limitations, particularly in video understanding where suitable pre-trained models are less common. Addressing this, our study introduces a novel cross-modality transfer learning approach from images to videos, which we call parameter-efficient image-to-video transfer learning. We present the Facial-Emotion Adapter (FE-Adapter), designed for efficient fine-tuning in video tasks. This adapter allows pre-trained image models, which traditionally lack temporal processing capabilities, to analyze dynamic video content efficiently. Notably, it uses about 15 times fewer parameters than previous methods, while improving accuracy. Our experiments in video emotion recognition demonstrate that the FE-Adapter can match or even surpass existing fine-tuning and video emotion models in both performance and efficiency. This breakthrough highlights the potential for cross-modality approaches in enhancing the capabilities of AI models, particularly in fields like video emotion analysis where the demand for efficiency and accuracy is constantly rising.

\end{abstract}

\section{INTRODUCTION}

Deep learning has risen to prominence as a transformative technology, delivering groundbreaking achievements in multiple fields. It has revolutionized tasks ranging from image recognition~\cite{resnet,vit}, emotion recognition~\cite{ma2021facial,aouayeb2021learning,canal2022survey} and action recognition~\cite{vivit}, pushing the limits of machine capabilities. These significant improvements in precision have altered the landscape of numerous industries, providing innovative solutions to problems that have persisted for years.

The rapid evolution of self-supervised models in machine learning, particularly in the domains of image~\cite{sun2023mae,radford2021learning} and language processing, has marked a significant advancement in the field. These models have demonstrated remarkable proficiency in a wide range of downstream tasks, setting new benchmarks across various applications. However, as these models grow in size, the traditional method of fully fine-tuning them for specific tasks becomes increasingly untenable~\cite{adapter} and computationally expensive~\cite{gowda2023watt}. This is predominantly due to the exorbitant computational costs and storage requirements associated with such processes.

Recognizing this challenge, the focus of recent research has shifted towards more efficient methods of transfer learning. The objective is to leverage the knowledge captured in these pre-trained models while minimizing the additional resources required for adaptation to new tasks~\cite{adapter,pan2022st,gowda2023optimizing}. Existing strategies predominantly concentrate on tasks within the same modality as the pre-trained model, such as image processing~\cite{liu2022clip,DFER-CLIP}. This approach, while effective, has inherent limitations, especially in modalities like video understanding where equivalent pre-trained models are either scarce or non-existent.

\begin{figure}[t]
    \centering    \includegraphics[width=0.99\linewidth]{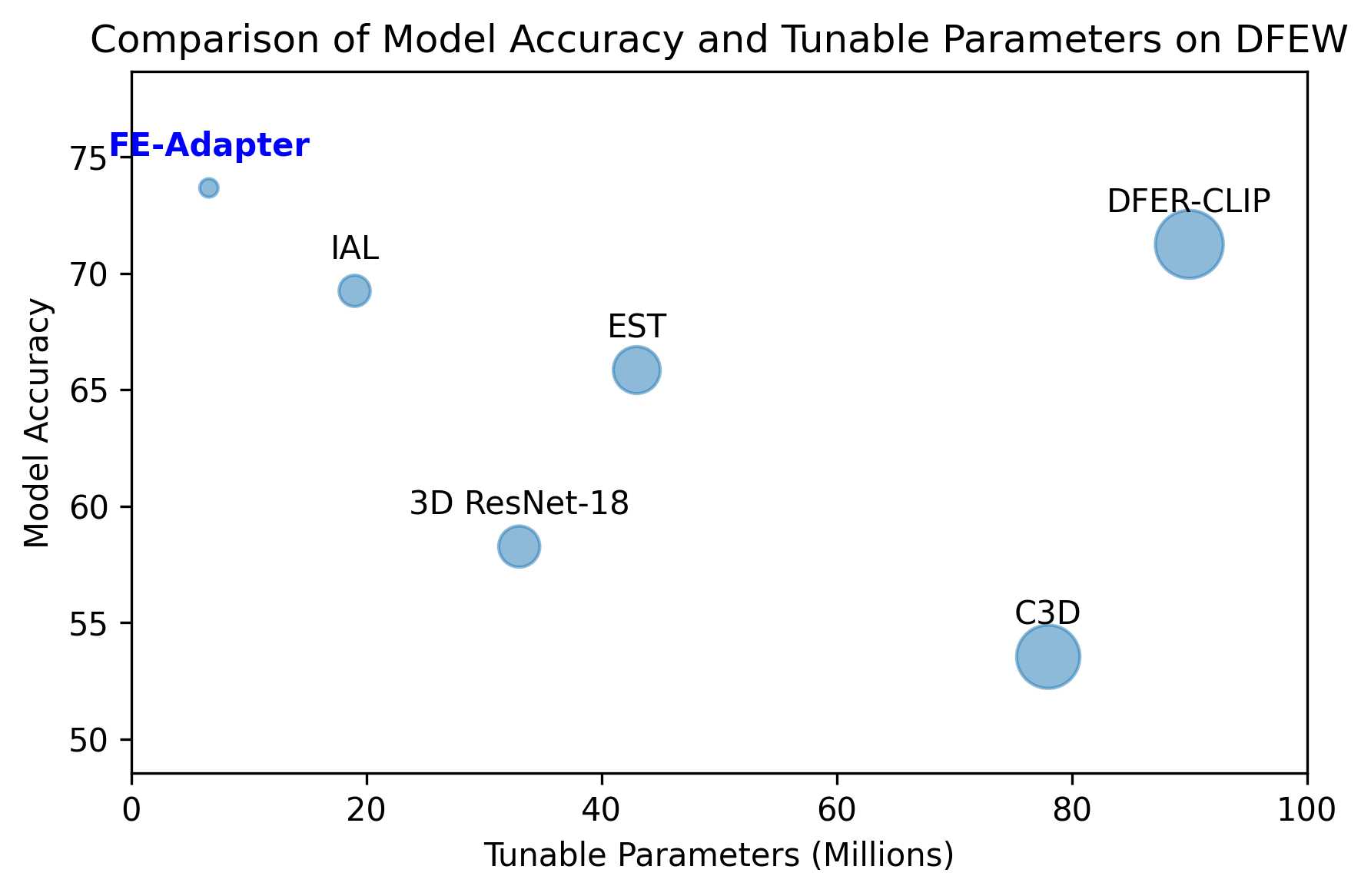}
\caption{A comparative analysis of various video-based models on the DFEW~\cite{dfew} dataset, showcasing the correlation between the number of tunable parameters (in millions) and model accuracy (\%). Our proposed FE-Adapter (highlighted in blue and bold), requires significantly fewer trainable parameters whilst outperforming recent SOTA models including vision-language models. The size of each bubble represents the number of tuneable parameters of the respective model. We compare with recent SOTA models such as IAL~\cite{IAL}, EST~\cite{EST}, and DFER-CLIP~\cite{DFER-CLIP}. We also compare against older 3D based models such as 3D-ResNet~\cite{resnet} and C3D~\cite{c3d}.}
    \label{fig:teaser}
\end{figure}

The introduction of vision transformers (ViTs)~\cite{vit} has resulted in significant improvements in general computer vision problems. However, ViTs often perform inferiorly compared to CNNs due to a lack of inductive bias and a tendency to focus on occlusions and noisy areas~\cite{aouayeb2021learning}. Specific approaches to improving ViTs for facial emotion recognition includes using attentional selective fusion~\cite{ma2021facial}, squeeze and excitation blocks~\cite{aouayeb2021learning} or neural resizing~\cite{hwang2022vision} among other approaches. Nonetheless, the application of ViTs in videos is very scarce.

In this context, our study introduces a novel concept in cross-modality transfer learning, explicitly transitioning from facial-image emotion understanding to video emotion understanding. We present the Facial-emotion Adapter (FE-Adapter), a bespoke solution designed for efficient fine-tuning in video-related tasks. The FE-Adapter stands out with its compact and effective architecture. It enables pre-trained image models, which inherently lack the capability to process temporal dynamics, to interpret and analyze video content effectively. Figure~\ref{fig:teaser} highlights the significant difference in tuneable parameters (size of bubble and x-axis) whilst improving upon the state-of-the-art.

One of the most compelling aspects of the FE-Adapter is its parameter efficiency. It requires approximately 8\% of the parameters per task (in comparison to current state-of-the-art models), translating to about 15 times fewer updated parameters than the state-of-the-art~\cite{DFER-CLIP}. This feature addresses the critical challenge of resource-intensive fine-tuning in large models. Through our extensive experiments in the realm of video emotion recognition, the FE-Adapter has demonstrated its ability to not only match but, in some instances, surpass both the traditional comprehensive fine-tuning approaches and the latest models dedicated to video emotion. This achievement is particularly noteworthy as it maintains the dual benefits of parameter efficiency and adaptability, making the FE-Adapter an invaluable tool in the field of video emotion recognition, where efficiency and flexibility are paramount.

\section{RELATED WORK}

\subsection{Emotion Recognition in Image} 
Emotion recognition in images has progressed significantly with the advent of deep learning. Convolutional Neural Networks (CNNs) have been extensively utilized~\cite{ko2018brief,li2021facial} for this task due to their ability to capture spatial hierarchies in facial features. 

Image classification models have been adapted for facial emotion recognition (FER), but challenges persist~\cite{canal2022survey} in natural environments due to issues like pose variations, occlusions, and distracting backgrounds. To tackle these, attention-based and region-based methods have emerged. Attention-based techniques focus on the most informative parts of facial expressions, while region-based strategies segment the face into sub-regions using various cropping techniques to minimize occlusions and background interference. 

However, they often overlook the interconnectivity between these regions. Vision Transformers (ViTs)~\cite{vit} have brought notable advancements in computer vision, but they sometimes fall short in comparison to CNNs, especially in handling occlusions and noisy areas. Methods to enhance ViTs for FER involve attentional selective fusion~\cite{ma2021facial}, squeeze and excitation blocks~\cite{aouayeb2021learning}, and neural resizing~\cite{hwang2022vision}. Yet, their application in video-based FER remains limited.

\subsection{Emotion Recognition in Videos} 
Video emotion recognition (VER) analyzes emotions in videos by combining temporal and spatial data. It differs from image recognition, using 3D CNNs~\cite{c3d,hara2017learning} to track spatial and temporal variations in facial expressions. 

VER faces challenges like pose changes, lighting, and obstructions, exacerbated by motion blur and varying frame rates. Solutions include a convolutional spatial transformer and a temporal transformer~\cite{zhao2021former}, handling occlusions and varied poses. Other approaches process videos in short clips~\cite{liu2022clip}, employing clip-based encoders and emotional intensity networks for emotion detection, or using intensity-aware losses~\cite{IAL} and Multi-Instance Learning~\cite{wang2023rethinking} to manage imprecise labels and capture time-based relationships. 

Self-supervised pre-training~\cite{sun2023mae} on large-scale data with a Transformer encoder is also used. DFER-CLIP~\cite{DFER-CLIP}, a visual-language model based on CLIP~\cite{radford2021learning}, specifically focuses on Dynamic Facial Expression Recognition (DFER), using a Transformer and textual descriptions to improve accuracy. While CLIP has been extensively adapted for video action recognition, its application in VER is less explored.

\subsection{Adapting Image models to Video}

The growing use of large pre-trained language models for various tasks has highlighted the importance of efficient tuning in NLP. This form of parameter-efficient tuning can be broadly categorized into two main approaches. 

First is the use of task-specific adapters~\cite{adapter,he2021towards}, which are light modules added between the layers of a pre-trained model. For efficiency, only these adapters are updated during task fine-tuning, leaving the bulk of the pre-trained model's parameters unchanged. 

The second approach is prompt tuning~\cite{liu2021p,lester2021power}, where learnable tokens are added either at the model's input or intermediate layers, and only these tokens are adjusted for each task whilst obtaining impressive results. 

While approaches have been proposed~\cite{pan2022st,gowda2023optimizing} to adapt image models to videos for tasks in video understanding, to the best of our knowledge nothing has been proposed for video emotion recognition. 

\section{METHODOLOGY}
\begin{figure}[t]
    \centering    \includegraphics[width=0.99\linewidth]{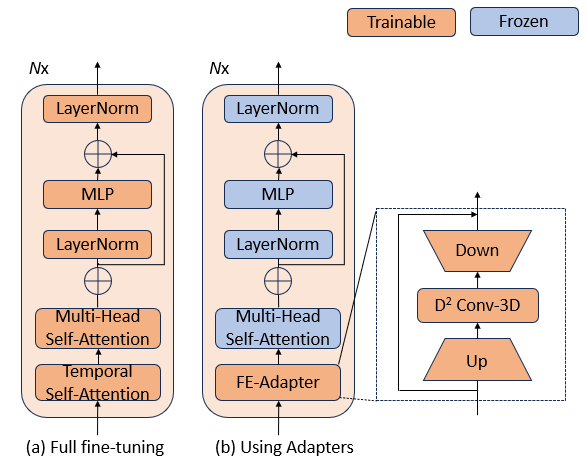}
\caption{Adapters ensure minimal parameter updates whilst keeping the generalization ability of the pre-trained model consistent. In comparison to (a) full fine-tuning, using (b) adapter modules significantly reduces tuneable parameter count.}
    \label{fig:adapters}
\end{figure}

\subsection{PRELIMINARIES}

To adapt highly-trained image models for video processing, it is essential to bridge the gap between still images and videos. We explore methods to aggregate image-based features for videos, with a focus on adapting Vision Transformers (ViTs) to video models like Timesformer~\cite{timesformer}.

In the simplest scenario, a pre-trained image model for video understanding involves aggregating frame features over time, such as through average pooling. Consider a video clip \( V \) defined as \( V \in \mathbb{R}^{T \times H \times W} \), where \( T \), \( H \), and \( W \) represent the number of frames, height, and width respectively. Each frame is divided into \( N = \frac{H \times W}{P^2} \) patches of size \( P \times P \). These patches are flattened and projected into patch tokens \( Z_t = [z_1, \ldots, z_s, \ldots, z_N] \), with \( z_s \in \mathbb{R}^d \) and \( d = 3 \times P^2 \), for \( t = 1, \ldots, T \). Positional embeddings are added to this series of feature vectors, along with a trainable class token. Each sequence of \( N + 1 \) tokens undergoes self-attention processing, retaining only the classification token \( z_{\text{cls}}^t \) for each sequence. Temporal average pooling is then applied to these class tokens\( z_{\text{final}} = \frac{1}{T} \sum_{t} z_{\text{cls}}^t \), to form a condensed representation for the entire clip. The final prediction is derived from a classifier using \( z_{\text{final}} \). This method, known as the Space-Only TimeSformer~\cite{timesformer}, averages spatial information over time.

Applying ViTs to the video domain involves enhancing spatio-temporal attention with additional temporal layers~\cite{vivit,timesformer,stmix}. While these models achieve higher accuracy, they typically require full fine-tuning for each task. This approach is inefficient in terms of parameters as it demands maintaining separate fully fine-tuned model parameters for every task.

\subsection{ADAPTERS}

Adapter modules, initially prominent in natural language processing (NLP), have been increasingly integrated into computer vision applications~\cite{adapter}. These modules are particularly known for their parameter efficiency, which means they add a relatively small number of trainable parameters to a pre-existing model to tailor it for specific tasks. This approach allows for adapting models to new tasks or datasets without the need to retrain the entire model, thus saving computational resources and time.

In the context of computer vision, adapter modules are inserted into pre-trained neural networks. These networks, designed for general tasks, are fine-tuned to perform specific vision tasks such as object detection, image segmentation, or image classification. The core idea behind adapters is to make these networks more flexible and efficient in learning task-specific features.

Given an input feature matrix \( \mathbf{X} \in \mathbb{R}^{N \times d} \) at the \(i\)-th layer of a neural network, the adaptation process performed by the adapter module can be formally defined as:
\begin{equation}
\text{Adapter}(\mathbf{X}) = \mathbf{X} + f(\mathbf{X}W_{\text{down}})W_{\text{up}},
\end{equation}
where \( W_{\text{down}} \in \mathbb{R}^{d \times r} \) denotes the down-projection layer, \( W_{\text{up}} \in \mathbb{R}^{r \times d} \) the up-projection layer, and \( f(\cdot) \) represents the activation function. The down-projection layer reduces the dimensionality of the input features \( \mathbf{X} \) to a lower-dimensional space \( \mathbb{R}^r \), aiming to capture the most relevant information for the task at hand. The up-projection layer then projects these features back to the original high-dimensional space \( \mathbb{R}^d \). This process is designed to enrich the original features with task-specific information. A residual connection adds the output of the adapter module to the original input features \( \mathbf{X} \), ensuring that the information present in the initial features is not lost, thereby conserving the input information.

The features \( \mathbf{X} \) referred to in this context are typically the embeddings or representations extracted from the \(i\)-th layer of the neural network. These embeddings encode the input data (e.g., images) into a form that the network can process to perform tasks like classification or detection. By applying adapter modules to these embeddings, the network can learn to adjust its representations to better suit specific tasks or datasets, enhancing performance with minimal additional trainable parameters.


\subsection{FE-ADAPTER}

The Adapter module, notably the ST-Adapter~\cite{pan2022st}, has been adapted for image models in video tasks. However, adapting facial image emotion models to videos remains unexplored. We propose using adapters as illustrated in Figure~\ref{fig:adapters}.

ST-Adapter uses depth-wise 3D-convolution for spatio-temporal analysis. This method might not be ideal for video emotion recognition due to its limitations in capturing complex spatial and temporal interactions, as it processes each channel independently, potentially overlooking emotional cues in facial expressions and body language. Depth-wise convolutions might not integrate spatial and temporal patterns effectively, which are crucial for emotion recognition.

We propose Dynamic Dilated Convolutions (D\(^2\)Conv3D)~\cite{d2conv3d} for enhanced performance in video emotion recognition tasks. The dynamic dilation rates of D\(^2\)Conv3D are well-suited for capturing intricate spatio-temporal patterns, especially in subtle facial emotions and temporal dynamics. This makes it more suitable for accurate emotion detection. The enhanced feature extraction capabilities of D\(^2\)Conv3D make it ideal for analyzing nuanced emotional cues in videos. We define our FE-Adapter as:
\begin{equation}
    \text{FE-Adapter}(X) = X + f\left(D^{2}\text{Conv3D}(XW_{\text{down}})\right)W_{\text{up}}
\end{equation}

Our Ablation Study highlights the importance of this choice for emotion recognition. In integrating the FE-Adapter for video emotion recognition, similar to adapters in NLP placed between Transformer layers~\cite{adapter}, various strategies are considered. Typically in NLP, adapters are positioned after both the Multi-Head Self-Attention (MHSA) and the Multi-Layer Perceptron (MLP). Some studies~\cite{adapter,pan2022st} suggest that a single adapter post-MLP is effective. 

For the FE-Adapter, we explore similar placements. Empirical evidence suggests that positioning a single FE-Adapter before the MHSA in each transformer block can enhance performance, allowing the FE-Adapter to preprocess input features effectively before the MHSA, and optimizing emotion recognition.

\section{EXPERIMENTAL ANALYSIS}

\subsection{DATASETS}

For our video emotion recognition task, we utilize a range of datasets, each offering unique characteristics and challenges. These include DFEW~\cite{dfew}, FERV39K~\cite{wang2022ferv39k}, and MAFW~\cite{liu2022mafw}. DFEW features 16,000 video clips across seven emotions, with 12,059 clips used for 5-fold cross-validation experiments. FERV39K provides 38,935 video sequences derived from four different scenarios, split into 31,088 for training and 7,847 for testing. MAFW, focusing on the video modality, comprises 10,045 video clips annotated with 11 emotions, using 9,172 clips in a 5-fold cross-validation setup. These diverse datasets allow for comprehensive training and evaluation of our emotion recognition models and also is commonly used in current state-of-the-art models.

\subsection{IMPLEMENTATION DETAILS}

In our experiments, we employ PF-ViT based on the ViT-B architecture, re-implemented with identical hyperparameters for AffectNet classification. We adapt this model for video tasks, using one adapter per PF-ViT block. The training involves the AdamW optimizer with a learning rate of 5e-4 and weight decay of 1e-2, alongside a batch size of 128. We process 16 frames per video, resized to 224×224. The learning rate is decayed using cosine annealing over 100 epochs. All experiments are conducted using an NVIDIA A100 GPU.

\subsection{ABLATION STUDY}

\subsubsection{Choice of Pre-Training Dataset} We use PF-ViT fine-tuned on AffectNet~\cite{mollahosseini2017affectnet} for adapting to video datasets. We also compare the use of RAF-DB~\cite{raf-db} and FERPLUS~\cite{ferplus}. These results are shown in Table~\ref{tab:pretraining_data}.

\begin{table}[]
\centering
\begin{tabular}{lcccccc}
\hline
Dataset & \multicolumn{2}{c}{DFEW~\cite{dfew}} & \multicolumn{2}{c}{FERV39k~\cite{wang2022ferv39k}} & \multicolumn{2}{c}{MAFW~\cite{liu2022mafw}}\\\
                  & UAR         & WAR        & UAR          & WAR    & UAR          & WAR     \\ \hline
RAF-DB            & 60.59      & 72.31     & 40.04       & 50.92   & 38.15 & 52.25    \\
FERPlus           & 60.66      & 72.42      & 40.44        & 51.31 & 38.87 & 53.48      \\
AffectNet       &  \textbf{60.89} & \textbf{73.67} & \textbf{41.44} & \textbf{52.11} & \textbf{39.41} &\textbf{55.02}   \\ \hline
\end{tabular}
\caption{Effect of Image models fine-tuned on different Image datasets.}
\label{tab:pretraining_data}
\end{table}

\subsubsection{Comparing Temporal Convolution Choices} 

In comparison to ST-Adapter, we use a D\(^2\)Conv3D instead of DW-Conv3D. Based on this, we compare the performance of different choices including a simple temporal aggregation of frame-by-frame features. We see a significant improvement using D\(^2\)Conv3D proving our hypothesis that discerns subtle facial emotions and temporal emotional dynamics. These results are in Table~\ref{tab:temp_conv}.

\begin{table}[]
\centering
\begin{tabular}{lcccccc}
\hline
Method & \multicolumn{2}{c}{DFEW~\cite{dfew}} & \multicolumn{2}{c}{FERV39k~\cite{wang2022ferv39k}} & \multicolumn{2}{c}{MAFW~\cite{liu2022mafw}}\\
                  & UAR         & WAR        & UAR          & WAR & UAR          & WAR        \\ \hline
 TA          & 53.66     & 68.51      & 31.92       & 46.41  & 31.14 & 47.51     \\
 Linear Probe & 53.85      & 69.66      & 33.12       & 47.22  & 34.59 & 50.17     \\
DW-Conv3D           & 57.65      & 71.15     & 37.69       & 50.40  & 37.11 & 51.91      \\
D\(^2\)Conv3D       &  \textbf{60.89} & \textbf{73.67} & \textbf{41.44} & \textbf{52.11} & \textbf{39.41} &\textbf{55.02}   \\ \hline
\end{tabular}
\caption{Comparing temporal convolution choices. `TA' refers to temporal aggregation.}
\label{tab:temp_conv}
\end{table}

\subsubsection{Adapter Position}

We consider two scenarios in terms of where we position the adapter. We first consider the global position i.e. what blocks do we place the adapter in. We do this by breaking the ViT-B to three sub-blocks such as 1-4, 5-8 and 9-12. We see that using an adapter in every block gives us best results. These results are in Table~\ref{tab:global}.

\begin{table}[]
\centering
\begin{tabular}{lcccccc}
\hline
1-4 & 5-8 & 9-12 & \multicolumn{2}{c}{FERV39k~\cite{wang2022ferv39k}} & \multicolumn{2}{c}{MAFW~\cite{liu2022mafw}} \\
     & &                    & UAR          & WAR    & UAR          & WAR      \\ \hline
  $\checkmark$    &     &    & 35.41 & 45.67 & 34.15 & 48.22  \\
  &  $\checkmark$   &    & 38.85 & 49.81 & 36.69 & 51.12   \\
   &  & $\checkmark$      & 40.44 & 51.08  & 38.22 & 53.65  \\
 & $\checkmark$ & $\checkmark$      & 40.89 & 51.66 & 39.00 & 54.11   \\
$\checkmark$ & $\checkmark$ & $\checkmark$      & \textbf{41.44} & \textbf{52.11} & \textbf{39.41} &\textbf{55.02}   \\ \hline
\end{tabular}
\caption{Comparing global adapter position choices.}
\label{tab:global}
\end{table}

We also consider where to place it within each block. With this we have a choice of placing it before the Multi-Head Self-Attention (MHSA) or after. We could also place it after the MLP. These results are shown in Table~\ref{tab:local}.

\begin{table}[]
\centering
\begin{tabular}{lcccc}
\hline
Position & \multicolumn{2}{c}{FERV39k~\cite{wang2022ferv39k}} & \multicolumn{2}{c}{MAFW~\cite{liu2022mafw}} \\
                        & UAR          & WAR     & UAR          & WAR     \\ \hline
After MLP & 41.35 & 51.89 &  39.25 & 54.78 \\
After MHSA & 41.39 & 52.07 & 39.33 & 54.92 \\
Before MHSA      & \textbf{41.44} & \textbf{52.11} & \textbf{39.41} &\textbf{55.02}   \\ \hline
\end{tabular}
\caption{Comparing local adapter position choices.}
\label{tab:local}
\end{table}

\subsection{COMPARISON TO STATE-OF-THE-ART}
We closely examine the performance of our proposed FE-Adapter against various leading models in the domain of facial emotion recognition. Table~\ref{tab:my_label} presents a comparative analysis of various methods in the context of facial emotion recognition using three popular benchmarks: DFEW, FERV39k, and MAFW. Each method is evaluated based on its performance in Unweighted Average Recall (UAR) and Weighted Average Recall (WAR) across these datasets. The table also includes the number of tunable parameters (in millions) for each method, indicating the complexity and potential computational demands.

\begin{table*}[]
    \centering
\begin{tabular}{|c|c|c|c|c|c|c|c|}
\hline \multirow{2}{*}{ Method } & \multirow{2}{*}{\begin{tabular}{c} 
Tunable Params (M)
\end{tabular}} & \multicolumn{2}{|c|}{ DFEW~\cite{dfew} } & \multicolumn{2}{|c|}{ FERV39k~\cite{wang2022ferv39k} } & \multicolumn{2}{|c|}{ MAFW~\cite{liu2022mafw} } \\
 & & UAR & WAR & UAR & WAR & UAR & WAR \\
\hline C3D~\cite{c3d} [ICCV'15] & 78 & 42.74 & 53.54 & 22.68 & 31.69 & 31.17 & 42.25 \\
\hline 3D ResNet-18~\cite{hara2017learning} [ICCVW'17] & 33 & 46.52 & 58.27 & 26.67 & 37.57 & - & - \\
\hline Former-DFER~\cite{zhao2021former} [MM'21] & 18 & 53.69 & 65.70 & 37.20 & 46.85 & - & - \\
\hline CEFLNet~\cite{liu2022clip} [IS'2022] & 13 & 51.14 & 65.35 & - & - & - & - \\

\hline IAL~\cite{IAL} [AAAI'23] & 19 & 55.71 & 69.24 & 35.82 & 48.54 & - & - \\
\hline EST~\cite{EST} [PR'23] & 43 & 53.43 & 65.85 & - & & - & - \\
\hline M3DFEL~\cite{wang2023rethinking} [CVPR'23] & - & 56.10 & 69.25 & 35.94 & 47.67 & - & - \\
\hline MAE-DFER~\cite{sun2023mae} [MM'23] & 85 & 63.41 & 74.43 & 43.12 & 52.07 & 41.62 & 54.31 \\
\hline DFER-CLIP~\cite{DFER-CLIP} [BMVC'23] & 90 & 59.61 & 71.25 & 41.27 & 51.65 & 38.89 & 52.55 \\
\hline \textbf{FE-Adapter (Ours)} & \textbf{6.6} & \textbf{60.89} & \textbf{73.67} & \textbf{41.44} & \textbf{52.11} & \textbf{39.41} & \textbf{55.02} \\
\hline
\end{tabular}
    \caption{Comparing FE-Adapter with state-of-the-art models using UAR (Unweighted Average Recall) and WAR (Weighted Average Recall) on three popular benchmarks. }
    \label{tab:my_label}
\end{table*}

Notable methods listed include C3D~\cite{c3d}, 3D ResNet-18~\cite{hara2017learning}, Former-DFER~\cite{zhao2021former}, CEFLNet~\cite{liu2022clip}, IAL~\cite{IAL}, EST~\cite{EST}, M3DFEL~\cite{wang2023rethinking}, MAE-DFER~\cite{sun2023mae} and DFER-CLIP~\cite{DFER-CLIP}. A key observation is the relatively low number of tunable parameters in the FE-Adapter method (6.6 million), which is significantly lower than most other methods. 

Overall, the table illustrates that FE-Adapter, despite its markedly lower complexity, competes well with or outperforms recent state-of-the-art models in facial emotion recognition across different benchmarks, highlighting its effectiveness and efficiency.

\section{ACKNOWLEDGEMENTS}

DAC was supported by the Pandemic Sciences Institute at the University of Oxford; the National Institute for Health Research (NIHR) Oxford Biomedical Research Centre (BRC); an NIHR Research Professorship; a Royal Academy of Engineering Research Chair; the Wellcome Trust funded VITAL project (grant 204904/Z/16/Z); the EPSRC (grant EP/W031744/1); and the InnoHK Hong Kong Centre for Cerebro-cardiovascular Engineering (COCHE).

\section{CONCLUSION}

In conclusion, this study presents an innovative cross-modality transfer learning approach, namely parameter-efficient image-to-video transfer learning, through the development of the Facial-Emotion Adapter (FE-Adapter). This novel approach addresses the challenges posed by the conventional method of fully fine-tuning large pre-trained models, which is resource-intensive in terms of training and storage. The FE-Adapter demonstrates the feasibility and effectiveness of using pre-trained image models for video understanding tasks. It does so by equipping these models with the capability to process dynamic video content, while significantly reducing the requirement for updated parameters – approximately 15 times fewer than previous methods. Our comprehensive experiments in video emotion recognition indicate that the FE-Adapter not only holds its own against but in some cases, outperforms both the traditional comprehensive fine-tuning methods and the latest models in this domain. Importantly, it achieves these results while upholding the advantages of parameter efficiency. This makes the FE-Adapter a highly valuable contribution to the field of video emotion recognition, offering a practical solution that combines efficiency with adaptability.

{\small
\bibliographystyle{ieee}
\bibliography{egbib}
}

\end{document}